\documentclass[10pt,twocolumn]{IEEEtran}
\usepackage{varwidth}
\usepackage{caption2}
\usepackage{graphicx,subfigure}
\usepackage{amsmath}
\usepackage{blindtext}
\usepackage{algorithm} 
\usepackage{algorithmic} 
\usepackage{epsfig}
\usepackage{graphics}
\graphicspath{{FIG/fig1/}{FIG/fig2/}{FIG/fig3/}{FIG/fig4/}{FIG/fig5/}{FIG/fig6/}{FIG/fig_pm/}}
\usepackage{color}
\usepackage{graphicx}
\usepackage{stfloats}
\usepackage{epsfig,amsfonts,multirow}
\usepackage{changebar}
\newcommand{\punt}[1]{}

\ifCLASSINFOpdf
\else

\fi

\begin{document}
\title{Elastic Net Hypergraph Learning for Image Clustering and Semi-supervised Classification }

\author{Qingshan~Liu,~\IEEEmembership{Seninor Member,~IEEE}, Yubao~Sun, Cantian~Wang, Tongliang~Liu and Dacheng~Tao,~\IEEEmembership{Fellow,~IEEE}
\thanks{}
\thanks{Q. Liu, Y. Sun and C. Wang are with the School of Information and Control, Nanjing University of Information Science and Technology, Nanjing 210044, China (e-mail: qsliu@nuist.edu.cn; sunyb@nuist.edu.cn; wangcantian0915@gmail.com).}
\thanks{T. Liu and D. Tao is with the Centre for Quantum Computation and Intelligent Systems and the Faculty of Engineering and Information Technology, University of Technology, Sydney, NSW2007, Australia (e-mail: tliang.liu@gmail.com; dacheng.tao@uts.edu.au).}}

\markboth{}%
{Shell \MakeLowercase{\textit{et al.}}: Elastic Net Hypergraph Learning with for Image Clustering}

\maketitle

\begin{abstract}
Graph model is emerging as a very effective tool for learning the complex structures and relationships hidden in data. Generally,
the critical purpose of graph-oriented learning algorithms is to construct an informative graph for image clustering
and classification tasks. In addition to the classical $K$-nearest-neighbor and $r$-neighborhood methods for graph
construction, $l_1$-graph and its variants are emerging methods for finding the neighboring samples of a center
datum, where the corresponding ingoing edge weights are simultaneously derived by the sparse reconstruction coefficients of
the remaining samples. However, the pair-wise links of $l_1$-graph are not capable of capturing the high order relationships
between the center datum and its prominent data in sparse reconstruction. Meanwhile, from the perspective of variable
selection, the $l_1$ norm sparse constraint, regarded as a LASSO model, tends to select only one datum from a group of data
that are highly correlated and ignore the others. To simultaneously cope with these drawbacks, we propose a new elastic
net hypergraph learning model, which consists of two steps. In the first step, the Robust Matrix Elastic Net model is
constructed to find the canonically related samples in a somewhat greedy way, achieving the grouping effect by adding
the $l_2$ penalty to the $l_1$ constraint. In the second step, hypergraph is used to represent the high order relationships
between each datum and its prominent samples by regarding them as a hyperedge. Subsequently, hypergraph Laplacian matrix
is constructed for further analysis. New hypergraph learning algorithms, including unsupervised clustering and
multi-class semi-supervised classification, are then derived. Extensive experiments on face and handwriting databases
demonstrate the effectiveness of the proposed method.
\end{abstract}


\IEEEpeerreviewmaketitle

\section{Introduction}

Graph model is widely regarded as an effective tool for representing the association relationships and intrinsic structures hiding in data.
Generally, graph model takes each data point as a vertex and links a pairwise edge to represent the association relationship
between two data points. In this way, data clustering is usually formulated as a graph partition problem without any assumption
on the form of the clusters \cite{shi2000normalized}, \cite{ng2002spectral}. Graph is also widely used as a basic tool
in many machine learning methods such as subspace learning \cite{sparsesubspace}, \cite{LRR}, manifold learning \cite{tenenbaum2000global},
\cite{roweis2000nonlinear}, \cite{belkin2003laplacian} and semi-supervised learning \cite{zhu2006semi}, \cite{zhu2003semi}.

\textbf{Related work:} How to construct an informative graph is a key issue in all graph-based learning methods.
The $K$-Nearest Neighbors (KNN) graph and $r$-neighborhood graph are two popular methods for graph construction. KNN connects each vertex
to its $k$-nearest neighbors, where $k$ is an integer number to control the local relationships of data. The $r$-neighborhood graph
connects each center vertex to the vertices falling inside a ball of radius $r$, where $r$ is a parameter that characterizes the local structure of data.
Although simple, these two methods have some disadvantages. For example, due to the use of uniform neighborhood size, they cannot produce datum-adaptive
neighborhoods that determine the graph structure, and thus they are unable to well capture the local distribution of data. To achieve better performance,
some similarity measurement functions, e.g., indicator function, Gaussian kernel function and cosine distance, are employed to encode the graph edge weights. However,
real-world data is often contaminated by noise and corruptions, and thereby the similarities estimated by directly measuring corrupted data may seriously deviate from
the ground truth.

Recently, Cheng \emph{et al.} \cite{cheng2010learning} proposed a robust and datum-adaptive method called $l_1$-graph, in which sparse representation
is introduced to graph construction. $l_1$-graph simultaneously determined both the neighboring samples of a datum and the corresponding edge weights by the sparse
reconstruction from the remaining samples, with the objective of minimizing the reconstruction error and the $l_1$ norm of the reconstruction coefficients.
Compared with the conventional graphs constructed by the KNN and $r$-neighborhood methods, the $l_{1}$-graph has some nice properties, e.g., the robustness to noise and the datum-adaptive ability.
Inspired by $l_{1}$-graph, a non-negative constraint is imposed on the sparse representation coefficients in \cite{he2011nonnegative}. Tang \emph{et al.} constructed a KNN-sparse graph for
image annotation by finding datum-wise one-vs-kNN sparse reconstructions of all samples \cite{tang2011image}. All these methods used multiple pair-wise edges (i.e., the non-zero prominent coefficients) to represent the relationships between the center datum and the prominent datums. However, the center datum has close
relationships with all the prominent datums, which is high-order rather than pair-wise. The pair-wise links in $l_1$-graph are not capable of capturing such high-order relationships, because some valuable
information may be lost by breaking a multivariant relationship into multiple pair-wise edge connections. In general, it is very crucial to establish
effective representations for these high-order relationships in image clustering and analysis tasks.

In terms of variable selection using linear regression model, the $l_1$ norm constrained sparse representation problem in $l_1$-graph can
be regarded as a LASSO problem \cite{las}, which takes the center datum as the response and the remaining data as the covariate predictors \cite{xiaotonggroup}.
According to the extensive studies in \cite{xiaotonggroup}, \cite{elasticnet}, the $l_1$ norm in LASSO has the shortcoming that each variable is
estimated independently and therefore the relationships and structures between the variables are not considered. More precisely, if there is a group of highly correlated variables,
then LASSO tends to select one variable from a group and ignore the others. In fact, it has been empirically observed that the prediction performance of LASSO is
dominated by the ridge regression if the high correlations between predictors existing \cite{elasticnet}. Intuitively, we expect that all the related data points are selected
as a group to predict the response. To this end, group sparsity techniques, e.g., the $l_{p,q}$ mixed norm, are suitable choices, because they favor the
selection of multiple correlated covariates to represent the response \cite{group}. However, the group sparsity regularization needs to know the grouping information.
In many cases, unfortunately, we are unaware of the grouping information.

\textbf{Motivation: }In contrast to pair-wise graph, a hypergraph is a generalization of a graph, where each edge (called hyperedge) is capable to connect
more than two vertices \cite{lovasz1974minimax}, \cite{zhou2006learning}. In other words, vertices with similar characteristics can all be enclosed by a hyperedge,
and thus the high order information of data, which is very useful for learning tasks, can be captured in an elegant fashion. Taking the clustering problem as an example,
it is often necessary to consider three or more data points together to determine whether they belong to the same cluster. As a consequence,
hypergraph is gaining much attention in these years. Agarwal \emph{et al.} \cite{agarwal2006higher}, \cite{agarwal2005beyond}
applied hypergraph for data clustering, in which clique average is performed to transform a hypergraph to a usual pair-wise graph. Zass and Shashua \cite{zass2008probabilistic}
adopted the hypergraph in image matching by using convex optimization. Hypergraph was applied to the problem of multilabel learning in \cite{sun2008hypergraph}
and video segmentation in \cite{huang2009video}. In \cite{tian2010integrative}, Tian \emph{et al.} proposed a semi-supervised learning method called HyperPrior
to classify gene expression data by using probe alignment as a constraint. \cite{zhou2006learning} presented the basic concept of hypergraph Laplacian and the
hypergraph Laplacian based learning algorithm. In \cite{huang2010image}, Huang \emph{et al.} formulated the task of image clustering as a problem of hypergraph
partition. In \cite{huang2011unsupervised}, a hypergraph ranking was designed for image retrieval. However, almost all the above methods use a simple KNN strategy
to construct the hyperedges. Namely, a hyperedge is generated from the neighborhood relationship between each sample and its $K$ nearest neighbors, which cannot
adaptively match the local data distribution. Hong \emph{et al.} integrated the idea of sparse representation to construct a semantic correlation hypergraph (SCHG) for image retrieval \cite{hong2012hypergraph},
which uses the top $K$ highest sparse coefficients to build a hyperedge. However, such a fixed order hyperedge still cannot adapt well to the local data distribution.
In addition, SCHG also adopted the $l_1$ norm as the sparsity measurement criterion, suffering the same shortcomings as LASSO and $l_1$-graph. In the nutshell,
the fundamental problem of an informative hypergraph model is how to define hyperedges to represent the complex relationship information, especially the group structure hidden in the data.

\textbf{Our Work:} In this paper, we propose a new elastic net hypergraph learning method for image clustering and semi-supervised classification.
Our algorithm consists of two steps. In the first step, we construct a robust matrix elastic net model by adding the $l_2$ penalty to the $l_1$ constraint
to achieve the group selection effect. The Least Angle Regression (LARS) \cite{las}, \cite{elasticnet} algorithm is used to find the canonically related
samples and obtain the representation coefficient matrix in a somewhat greedy way, unlike the convex optimization algorithms adopted in \cite{cheng2010learning} and \cite{sparsesubspace}.
In the second step, based on the obtained reconstruction, hyperedge is used to represent the high-order relationship between a datum and its prominent
reconstruction samples in the elastic net, resulting in an elastic net hypergraph. A hypergraph Laplacian matrix is then constructed to find the spectrum
signature and geometric structure of the data set for subsequent analysis. Compared to previous works, the proposed method can both achieve grouped selection
and capture high-order group information of the data by elastic net hypergraph. Lastly, new hypergraph learning algorithms, including unsupervised and semi-supervised
learning, are derived based on the elastic net hypergraph. Experiments on the Extended Yale B, the PIE face databases and the USPS handwriting database demonstrate the effectiveness
of the proposed method. 
The main innovations of our paper are summarized below:
\begin{itemize}
  \item  Robust Matrix Elastic Net is designed to find the canonical groups of predictors from the dictionary to reconstruct the response sample. More specially,
if there is a group of samples among which the mutual correlations are very high, our model tends to recognize them as a group and automatically include the whole group
into the model once one of its sample is selected (¡®group selection¡¯), which is very helpful for further analysis.
  \item In order to link a sample with its selected groups of predictors, an elastic net hypergraph model, instead of the traditional pair-wise graph,
is proposed, where a hyperedge represents the high-order relationship between one datum and its prominent reconstruction samples in the elastic net.
This paper devotes to construct an informative hypergraph for image analysis. Our model can effectively
represent the complex relationship information, especially the group structure hidden in the data, which is beneficial for clustering and
semi-supervised learning derived upon the constructed elastic net hypergraph.
\end{itemize}

In the following sections, we will first introduce the preliminaries of hypergraph. Section III details the construction of Elastic Net Hypergraph. Section V presents
the clustering and semi-supervised learning defined on the constructed hypergraph model. Experimental results and analysis are given in Section IV and Section VI concludes the paper.

\section{Hypergraph Preliminaries}
Assuming $V$ represents a finite set of samples, and $E$ is a family of hyperedge $e$ of $V$ such that $\bigcup_{e\in E} = V$,
A positive number $w(e)$ is associated with each hyperedge $e$, called the weight of hyperedge $e$. $G=(V,E,W)$ is then called a weighted hypergraph
with the vertex set $V$, the hyperedge set $E$ and the weight matrix $W$. An incidence matrix $H$ (of size $|V|\times|E|$) denotes the relationship between the vertices and the hyperedges, with entries defined as:
\begin{equation}\label{7}
  h(v_i,e_j) = \begin{cases}
    1,\quad if \quad v_i \in e_j \\
    0,\quad otherwise.
\end{cases}
\end{equation}
That is, $H$ indicates to which hyperedge a vertex belongs. Based on $H$, the vertex degree of each vertex $v_i \in V$ and the edge degree of hyperedge $e_j \in E$ can be calculated as:
\begin{equation}\label{7}
d(v_i)=\sum_{e_j \in E}w(e_j)h(v_i,e_j),
\end{equation}

\begin{equation}\label{7}
\delta(e_j)=\sum_{v_i \in V}h(v_i,e_j).
\end{equation}
For convenience, Table \ref{natations} lists the important notations used in the rest of this paper.
\begin{table}
  \centering
\caption{Important hypergraph natations used in the paper and their descriptions }\label{natations}
  \small\begin{tabular}{|c|p{6cm}|}
     \hline
  Notation &   Description \\
\hline
     $G = (V,E,W)$ & The representation of a hypergraph with the vertex set $V$, the hyperedge set $E$, and the hyperedge weight matrix $W$ \\
\hline
     $u,v$ & Vertices in the hypergraph \\
\hline
     $D_{v}$ & The diagonal matrix of the vertex degrees \\
\hline
     $D_{e}$ & The diagonal matrix of the hyperedge degrees \\
\hline
     $H$ & The incidence matrix for the hypergraph \\
\hline
     $W$ & The diagonal weight matrix and its$(i,i)$-th element is the weight $w(e_{i})$ of the $i$-th hyperedge $e_i$ \\
\hline
     $L$ & The constructed hypergraph Laplacian matrix \\
\hline
     $d(v_i)$ & The degree of the vertex $v_i$ \\
\hline
     $\delta(e_i)$ & The degree of the hyperedge $e_i$ \\
\hline
     $w(e_i)$ & The weight of the hyperedge $e_i$ \\
     \hline
   \end{tabular}
\end{table}

\begin{figure}
\begin{center}
\includegraphics[scale=0.35]{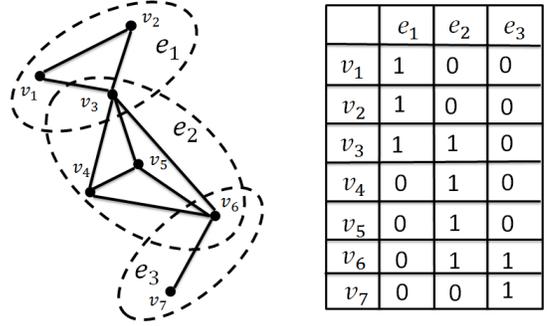} 
\end{center}
\caption{An example of hypergraph (left) and its corresponding $H$ matrix (right). Each hyperedge is marked by an ellipse.}
\label{hypergraphexample}
\end{figure}

From the above definition, the main difference between a hypergraph and a pair-wise graph (For convenience, we call it simple graph in the following) lies in that a hyperedge can link more than two vertices.
Thus, the hypergraph acts as a good model to represent local group information and complex relationship between samples.
Taking Fig. \ref{hypergraphexample} as an example, there are seven data points, and they are attributed to three local groups.
One may construct a simple graph, in which two vertices are joined together by an edge if they are similar. However, simple graph
cannot represent the group information well due to its pair-wise links. Different from a simple graph, a hypergraph can enclose a local
group as one hyperedge according to the $H$ matrix shown in the right of Fig. \ref{hypergraphexample}. Thus, the constructed hypergraph
is able to represent the local group information hidden in the data.

\section{ Elastic Net Hypergraph }

The hypergraph has been proposed as a natural way to encode higher order relationships in unsupervised and semi-supervised learning. By enclosing all the vertices
with common attributes or close relationships within one hyperedge, we can effectively describe the high-order information of the data. Nevertheless, how to discover
the related samples to form hyperedges and compute their weights is the key issue in the hypergraph construction. Most previous works have adopted the KNN method to
generate the hyperedge, whereby each sample and its $K$ nearest neighbors form a hyperedge \cite{huang2010image}, \cite{huang2011unsupervised}. The method is very simple,
but it is not adaptive to local data distribution and some inherent information may be lost in the construction of hypergraphs.

In this section, we propose a process for constructing the so-called elastic net hypergraph (ENHG), in which each sample acts as a vertex and the hyperedge associated
with each sample describes its robust elastic net driven reconstruction from the remaining samples. A robust matrix elastic net
model is first designed for discovering the group structures and relationships hidden in the data. For each data point, we find the canonically related samples from the remaining samples
to reconstruct it by the elastic net model. We then use the non-zero prominent elements of the representation to adaptively seek the most relevant neighbors to form a hyperedge,
so that the data points in that hyperedge have strong dependencies. By regarding the elastic net representation of each data point as a feature, we compute the hyperedge
weight by the sum of the mutual affinity between two data points calculated by the dot product between two features. The details are presented in the following sub-section.

\subsection{Robust Matrix Elastic Net for Group Representation}
For a general data clustering or classification problem, the training sample set is assumed to be stacked as a matrix $X =[x_1,x_2,\ldots,x_n]\in\mathbb{R}^{d \times n}$,
whose columns are $n$ data points drawn from $d$ dimensional feature space. In practice, the data points $X$ may be contaminated by gross error $S$,
\begin{equation}\label{dataandnoise}
  X=X_0+S,
\end{equation}
where $X_0$ and $X$ represent the clean data and the observed data respectively, $S=[s_1,s_2,\ldots,s_n]\in\mathbb{R}^{d \times n}$ is the error matrix. The $i$-th sample
is contaminated by error $s_i$, which can present as noise, missed entries, outliers and corruption. Then the clean data $X_0$ can be represented by a linear combination
of atoms from the dictionary $A =[a_1,a_2,\ldots,a_m]\in\mathbb{R}^{d \times m}$ ($m$ is the atom number of $A$) as:
\begin{equation}\label{matrixrep}
  X = AZ+S,
\end{equation}
where $Z =[z_1,z_2,\ldots,z_n]\in\mathbb{R}^{m \times n}$ is the coefficient matrix, and $z_i$ is the representation of $x_i$ upon the dictionary $A$. The dictionary $A$
is often redundant and over-complete. Hence there can be many feasible solutions to problem (\ref{matrixrep}). A popular method is to impose the common $l_1$ sparsity
criteria, known as sparse linear representation. Intuitively, the sparsity of the coding coefficient vector can be measured by the $l_{0}$ norm to count the nonzero coefficients
in the representation. It has been shown that under certain conditions, the $l_1$ norm optimization can provide us the sparse solution with similar nonzero supports as the
$l_0$ norm optimization \cite{donoho}.

From the view of variable selection, the sparse linear representation problem can be cast as a problem of sparse covariate selection via a linear regression model by taking
the dictionary matrix $A$ as an observation of the covariate and the query matrix $X$ as the response \cite{xiaotonggroup}. The $l_1$ norm constrained sparse linear representation
can be regarded as a LASSO model, which seeks to predict an output by linearly combining a small subset of the features that describe the data. As a result of efficient optimization
algorithms and the well-developed theory for generalization properties and variable selection consistency, the $l_1$ norm regularization has become a popular tool for variable selection
and model estimation. However, the $l_1$ norm has its shortcomings in that each variable is estimated independently, regardless of its position in the input feature vector. If there
is a group of variables among which the pair-wise correlations are very high, then LASSO tends to select only one variable from the group and does not care which one is selected.
It lacks the ability to reveal the grouping information. It has been empirically observed that if there are high correlations between predictors, the prediction performance of
LASSO is dominated by ridge regression. To overcome these limitations, the elastic net adds a quadratic part to the $l_1$ regularization, which can be regarded as a combination of
LASSO and ridge regression. Here we take sample-specific corruption as an example, $S$ indicates the phenomenon that a fraction of the data points (i.e., columns $x_i$ of the data
matrix $X$) is contaminated by a large error. By using the sample set $X$ itself as the dictionary, the matrix elastic net is modeled by
\begin{equation}\label{RoustMEN}
\begin{array}{*{20}{l}}
{\mathop {\min }\limits_{Z,S} {{\left\| Z \right\|}_1} + \lambda \left\| Z \right\|_F^2 + \gamma {{\left\| S \right\|}_{2,1}}}\\
{s.t.\;X = XZ + S,\;diag(Z)=0,}
\end{array}
\end{equation}
where the ``entrywise" $l_1$ norm of the matrix $Z$ is defined by ${\left\| Z \right\|_1}= \sum\limits_{i = 1}^m {\sum\limits_{j = 1}^n {\left| {{z_{i,j}}} \right|}}$, ${\left\| Z \right\|_F}$
is the Frobenius norm of the matrix $Z$, ${\left\|  \cdot  \right\|_{2,1}}$ denotes the ${l_{2,1}}$ mixed norm for dealing with sample-specific corruptions, computed as the sum of the
$\ell_2$ norm of the columns of the matrix: ${\left\| S \right\|_{2,1}} = \sum\limits_{j = 1}^n {{{\left\| {{s_j}} \right\|}_2}}$, $\lambda$ is the weight parameter of the quadratic
part and $\gamma$ is the regularization parameter to trade off the proportion $XZ$ and $S$. An additional constraint $diag(Z)=0$ is introduced, which is used to avoid the trivial
solution of representing a point as a linear combination of itself. In other words, each datum is reconstructed by the linear combination of the remaining samples, which can be
used to discover the group structures and relationships hidden in the data. The elastic net regularization encourages the grouping effect, favoring the selection of multiple correlated
data points to represent the test sample.

Now we start out to solve the model (\ref{RoustMEN}). First, by replacing $S$ with $X-XZ$, we can transform Eq. (\ref{RoustMEN}) into the following equivalent equation,
\begin{equation}\label{RoustMEN-EQ}
\begin{array}{*{20}{l}}
{\mathop {\min }\limits_{Z} {{\left\| Z \right\|}_1} + \lambda \left\| Z \right\|_F^2 + \gamma {{\left\| X-XZ \right\|}_{2,1}}}\\
{s.t.\;diag(Z)=0.}
\end{array}
\end{equation}
This objective function is to obtain the elastic net decomposition of all the samples, which can be indeed solved in a column-by-column fashion.
Namely, it is equivalent to solve the elastic net decomposition $z_i$ of each sample $x_i$ respectively. Inspired by \cite{cheng2010learning}, we cope with the constraint $z_{i,i}=0$
by eliminating the sample $x_i$ from the sample matrix $X$ and the elastic net decomposition of sample $x_i$ can be formulated as,
\begin{equation}\label{subproblem-z}
{\mathop {\min }\limits_{{z'_i}} {{\left\| {{z'_i}} \right\|}_1} + \lambda \left\| {{z'_i}} \right\|_2^2}+\gamma {\rm{ }}{\left\| {{x_i} - {B_i}{{z'}_i}} \right\|_2},\\
\end{equation}
where the dictionary matrix ${B_i} = [{x_1},{x_2},...,{x_{i - 1}},{x_{i + 1}},....,{x_n}] \in {R^{d \times (n - 1)}}$ and the decomposition coefficient ${z'_i} \in {R^{n - 1}}$.
It can be found that Eq. (\ref{subproblem-z}) is a typical elastic net model as in \cite{elasticnet}. Thus, we directly adopt the LARS-EN algorithm \cite{elasticnet},\cite{las} to
solve Eq. (\ref{subproblem-z}), which can compute the entire elastic net regularization paths with the computational effort of a single ordinal least squares fit. Since Eq. (\ref{subproblem-z})
is a convex problem, LARS-EN has been proved to converge to the global minimizer. After all the samples have been processed, the coefficient matrix can then be augmented as $n\times n$ dimensional matrix
by adding zero to the diagonal elements. Finally, we can obtain the coefficient matrix $Z$ and the clean data $X_0=XZ$ from the given observation matrix $X$, the gross error $S$ can be
accordingly computed as $X-XZ$. In terms of the reconstruction relationship of each vertex, we can define the hyperedge as the current vertex and its reconstruction, and predict the cluster
or label information through the hypergraph defined on the obtained elastic net representation.

\subsection{Hyperedge construction} Given the data, each sample $x_i$ forms a vertex of the hypergraph $G$, and can be represented by the other samples as in Eq. (\ref{RoustMEN}),
where $z_i$ is its sparse coefficients, naturally characterizing the importance of the other samples for the reconstruction of $x_i$. Such information is useful for recovering the
clustering relationships among the samples. Although there are many zero components in $z_i$, sample $x_i$ is mainly associated with only a few samples with prominent non-zero
coefficients in its reconstruction. Thus, we design a quantitative rule to select the prominent samples and define the incidence matrix $H$ of an ENHR as:
\begin{equation}\label{hyperedge}
  h(v_{i},e_{j}) = \begin{cases}
    1,\quad if\; \left| {{z_{ij}}} \right| > \theta  \\
    0,\quad otherwise,
\end{cases}
\end{equation}
where $\theta$ is a small threshold. For example, $\theta$ can be set as the mean values of $\left| z_i \right|$. It can be seen that a vertex $v_i$ is assigned to $e_j$ based on
whether the reconstruction coefficients $z_{ij}$ is greater than the threshold $\theta$. We take each sample as a centroid and form a hyperedge by the centroid and the selected
most relevant samples in the elastic net reconstruction. The number of neighbors selected by Eq. (\ref{hyperedge}) is adaptive to each datum, which is be propitious to capture the
local grouping information of non-stationary data.

\subsection{Computation of hyperedge weights} The hyperedge weight also plays an important role in the hypergraph model. In \cite{hong2012hypergraph}, the non-zero coefficients are
directly taken to measure the pair-wise similarity between two samples in the hyperedge. This is unreasonable, because the non-zero coefficients naturally represent the reconstruction
relationship, but not the explicit degree of similarity. In this paper, we take each sparse representation vector $z_i$ as the sparse feature of $x_i$, and we measure the similarity
between two samples by the dot product of two sparse vectors as

\begin{equation}\label{7}
  M(i,j) = \left| {\langle {z_i},{z_j}\rangle } \right|.
\end{equation}
The affinity matrix can be calculated as: $M = \left| {{Z^T}Z} \right|$, and the hyperedge weight $w(e_i)$ is computed as follows:
\begin{equation}\label{7}
 w(e_i) = \sum_{v_{j}\in e_{i},j \ne i}h(v_{j},e_{i})M(i,j).
\end{equation}

Based on this definition, the compact hyperedge (local group) with higher inner group similarities is assigned a higher weight, and a weighted hypergraph $G = (V,E,W)$ is subsequently
constructed.
The ENHG model construction is summarized in \textbf{Algorithm 1}.

\begin{algorithm}[htb]
\renewcommand{\algorithmicrequire}{\textbf{Input:}}
\renewcommand\algorithmicensure {\textbf{Procedure:}}
\caption{ The process of constructing elastic net hypergraph (ENHG)}
\label{alg:Framwork}
\begin{algorithmic}[1]
\REQUIRE ~~\\
Data matrix $X =[x_1,x_2,\ldots,x_n]\in\mathbb{R}^{d \times n}$, regularized parameters $\lambda$, $\gamma$ and threshold $\theta$.
\ENSURE ~~\\
\STATE Normalize all the samples to zero mean and unit length.
\STATE Solve the following problem to obtain the optimal solution $Z$:\\
$ $ $ $ $ $ $ $ $ $ $ $ $ $ $ $ $ $ $ $ $ $ $ $ $ $ $\begin{array}{l}
\mathop {\min }\limits_{Z,S} {\left\| Z \right\|_1} + {\lambda}\left\| Z \right\|_F^2 + \gamma {\left\| S \right\|_{2,1}}\\
s.t.\;X = XZ + S,\;diag(Z)=0.
\end{array}$
\STATE The incidence matrix $H$ of an ENHG can be obtained based on the reconstruction coefficients $Z$: \\
$ $ $ $ $ $ $ $ $ $ $ $ $ $ $ $ $ $ $ $ $ $ $ $ $ $ $ $
$h(v_{i},e_{j}) = \begin{cases}
    1,\quad if\;z_{ij}> \theta \\
    0,\quad otherwise.
\end{cases}$
\STATE The affinity matrix can be derived by the similarity relationship from the reconstruction coefficients: \\
$ $ $ $ $ $ $ $ $ $ $ $ $ $ $ $ $ $ $ $ $ $ $ $ $ $ $ $ $ $ $ $ $ $ $ $ $ $ $ $  $M(i,j) = \langle z_{i},z_{j}\rangle$.
\STATE Compute the hyperedge weight $w(e_i)$ by \\
$ $ $ $ $ $ $ $ $ $ $ $ $ $ $ $ $ $ $ $ $ $ $ $  $w(e_i) = \sum_{v_{j}\in e_{i}}h(v_{j},e_{i})M(i,j)$.
\RETURN  The incidence matrix $H$ and the hyperedge weight matrix $W$ of ENHG.
\end{algorithmic}
\end{algorithm}

\section{ Learning with Elastic net Hypergraph }
A well-designed graph is critical for those graph-oriented learning algorithms. In this section, we briefly introduce how to benefit from ENHG
for clustering and classification tasks. Based on the proposed ENHG model, a hypergraph Laplacian matrix is constructed to find the spectrum signature
and geometric structure of the data set for subsequent image analysis. Then, we formulate two learning tasks, i.e., spectral clustering and semi-supervised classification for
image analysis formulated in terms of operations on our elastic net hypergraph. The principal idea is to perform spectral decomposition on the Laplacian matrix of the hypergraph
model to obtain its eigenvectors and the eigenvalues \cite{zhou2006learning}. Our elastic net hypergraph Laplacian matrix is also computed as
\begin{equation}\label{Laplacian}
 L = I-D_v^{-1/2}HWD_e^{-1}H^T D_v^{-1/2},
\end{equation}
where $D_v$ and $D_e$ are the diagonal matrix of the vertex degrees and the hyperedge degrees, respectively. Based on the elastic net hypergraph model and its Laplacian matrix,
we can design different learning algorithms.
\subsection{Hypergraph spectral clustering}
Clustering, or partitioning similar items into dissimilar groups, is widely used in data analysis and is applied in various areas such as, statistics, computer science, biology
and social sciences. Spectral clustering is a popular algorithm for this task and is a powerful technique for partitioning simple graphs. Following \cite{zhou2006learning}, we
develop an ENHR-based spectral clustering method.
The main steps of spectral clustering based on ENHG are as follows:
\begin{enumerate}
  \item Calculate the normalized hypergraph Laplacian matrix by Eq. (\ref{Laplacian}).
  \item Calculate the eigenvectors of L corresponding to the first k eigenvalues (sorted ascendingly), denoting the eigenvectors by $C = \left[c_{1}, c_{2}, \ldots, c_{k}\right]$.
  \item Denote the $i$-th row of $C$ by $y_{i}\left(i=1, \ldots ,n\right)$, clustering the points $(y_{i} )_{i=1, \ldots ,n}$ in $\mathbb{R}^k$ with K-Means algorithm into
clusters $c_{1}, c_{2}, \ldots, c_{k}$.
  \item Finally, assign $x_i$ to cluster $j$ if the $i$th row of the matrix $C$ is assigned to cluster $j$.
\end{enumerate}

\subsection{Hypergraph Semi-supervised classification }
Now we consider semi-supervised learning on ENHG. Given an ENHG model $G=(V, E, W)$, each vertex $v_{i(1 \le i \le n)}$ represents a data point, $n$ is the total number of samples/vertices. Partial samples are labeled as $y_i$ from
a label set $L = \{ 1,...,c\}$; $c$ is the total number of categories and the remaining samples are unlabeled. The goal of hypergraph semi-supervised
learning is to predict the labels of the unlabeled samples according to the geometric structure of the hypergraph \cite{zhou2006learning}, \cite{zhou2004learning}, \cite{zhou2005learning}.
Due to the strong similarity of the data in a hyperedge, we try to assign the same label to all the vertices contained in the same hyperedge, and it is then straightforward to derive
a semi-supervised prediction from a clustering scheme. Define a $n \times c$ non-negative matrix $F = [{F_1};{F_2};...,{F_n}]$ corresponding to a classification on the $G$ by labeling
each vertex $v_i$ with a label ${y_i} = \arg {\max _{1 \le j \le c}}{F_{ij}}$. We can understand $F$ as a vectorial classification function $f:V \to {R^c}$, which assigns a label
vector $f(v)$ to a vertex $v \in V$.

The hypergraph semi-supervised learning model can be formulated as the following regularization problem,
\begin{equation}\label{7}
 \arg\min_{F}R_{emp}(F) + \lambda\Omega(F),
\end{equation}
where $\Omega(F)$ is a regularizer on the hypergraph, $R_{emp}(F)$ is an empirical loss, and $\lambda>0$ is the regularization parameter.
The regularizer $\Omega(F)$ on the hypergraph is defined by
\begin{equation}\label{hyperreg}
\begin{split}
 \Omega(F) = \frac{1}{2}\sum_{e\in E}\sum_{u,v\in e}\frac{w(e)H(u,e)H(v,e)}{\delta(e)}\\\times
\left(\frac{f(u)}{\sqrt{d(u)}}-\frac{f(v)}{\sqrt{d(v)}}\right)^{2}=Tr({F^T}LF),
\end{split}
\end{equation}
where $Tr$ is the matrix trace, and $L$ is the normalized hypergraph Laplacian matrix. Eq. (\ref{hyperreg}) measures how smoothly the classification function defined
on these points (vertices) changes with respect to their neighborhoods within the hyperedge. For the empirical loss, we define an $n \times c$ matrix $Y$ with ${Y_{ij}} = 1$ if $v_i$
is labeled as $y_{j}=j$ and ${Y_{ij}}=0$ otherwise. Note that $Y$ is consistent with the initial labels assigned according to the decision rule. To force the assigned labels to
approach the initial labeling $Y$, the empirical loss can be defined as follows:
\begin{equation}\label{8}
 R_{emp}(F) = \left\| {F - Y} \right\|_F^2 = \sum_{v_{i}\in V}\left(f(v_{i})-Y_{i}\right)^{2}.
\end{equation}

Differentiating the regularization framework with respect to $F$, we can obtain a linear system for achieving the classification matrix $F$. With the least square loss function,
as shown in \cite{zhou2006learning}, the classification matrix $F$ can be directly given by $F = {(I - \alpha \Theta )^{ - 1}}Y$ with iterations,
where $\Theta = D_v^{-1/2}HWD_e^{-1}H^T D_v^{-1/2}$, $\alpha$ is a parameter in (0, 1). The predicted label for each point $v_i$ is determined using:
\begin{equation}\label{semi-classificatin}
  {y_i} = \arg {\max _{1 \le j \le c}}{F_{ij}}.
\end{equation}
\subsection{Discussions }
\begin{figure}[!t] 
\centering
\includegraphics[scale=0.60]{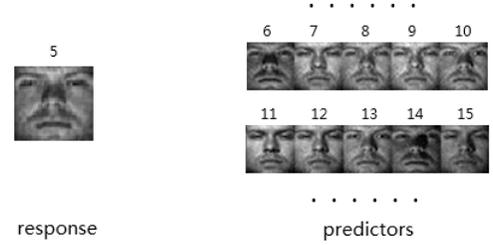} 
\caption{The response image (left) and the 6th to 15th predictor images (right). }
\label{responseimage}
\end{figure}
\begin{figure}[!t] 
\centering
\includegraphics[scale=0.47]{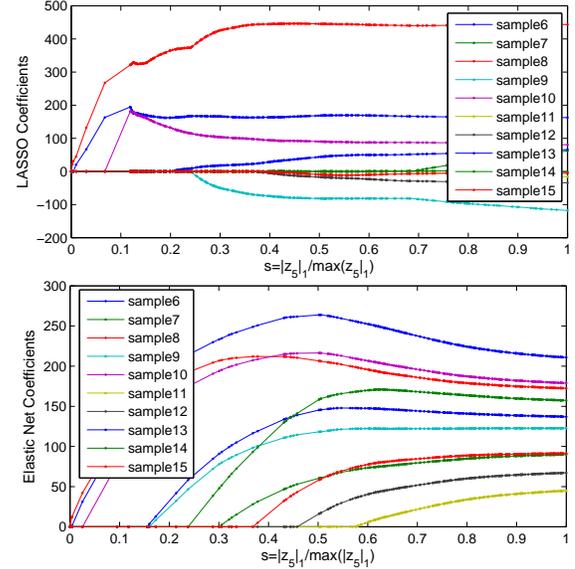} 
\caption{Comparison between the LASSO and elastic net variables selection path as a function of $s = \frac{{{{\left| {{z_5}} \right|}_1}}}{{\max ({{\left| {{z_5}} \right|}_1})}}$, among which ${z_5}$ represents the elastic net coefficient of the fifth sample and $\max ({\left| {{z_5}} \right|_1})$ means the max of the $l_1$ norm of coefficients in the fifth sample's solution path. }
\label{solutionpath}
\end{figure}
\begin{figure}[!t]
\centering
\includegraphics[scale=0.6]{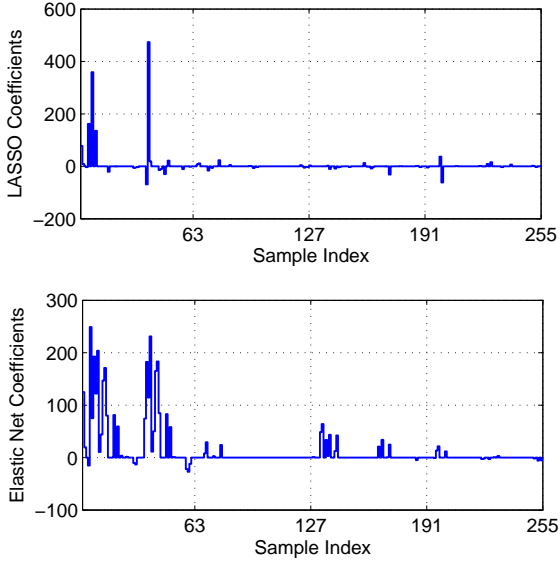}%
\caption{ The reconstruction coefficients of the fifth face image of the first individual using the LASSO model and our elastic net model.}
\label{elasticnetcoef}
\end{figure}
Based on the above description, we can find that the computation of the Laplacian matrix plays a key role in the two learning
algorithms. Here, we discuss the construction of our elastic net hypergraph Laplacian matrix, specifically the hyperedge construction, through a
series of experiments. The quantitative results of the two learning algorithms upon the face and handwritten digits databases and comparison with other
algorithms will be presented in Section V.

The authors of \cite{elasticnet} have argued that the elastic net promotes the group selection of canonically related samples. Qualitatively speaking,
a regression method exhibits the grouping effect if the regression coefficients of a group of highly correlated variables tend to be equal (up to a change
of sign if negatively correlated). Theorem 1 of \cite{elasticnet} pointed out the quantitative relationship between the consistency of sample $x_i$'s
and $x_j$'s coefficient paths and their correlation $\rho = x_i^T{x_j}$. To empirically inspect the group selection effect of our elastic net model, we perform
a number of evaluation experiments on the Extended Yale Face Database B \cite{lee2005acquiring} and examine the consistency of the solution path.
We select the first four individuals as the sample set $X$. Each individual has 64 near frontal images under different illuminations. We take each
sample as a vertex, so the hypergraph size is equal to the number of training samples, and $X$ is the sample matrix. The evaluation experiment on
the fifth face image of the first individual is presented for illustration. The response image (the fifth image) and partial predictor images (6th to 15th)
are shown in Fig. \ref{responseimage}. 

Fig. \ref{solutionpath} compares the solution path of the fifth face image of the first individual (response) in our elastic net model and the LASSO model.
The coefficient paths of the sixth to fifteenth samples (predictor) in the LASSO and the elastic net model are displayed. We adopt $s = \frac{{{{\left| {{z_5}} \right|}_1}}}{{\max ({{\left| {{z_5}} \right|}_1})}}$
as the horizontal axis. The vertical axis represents the coefficients value of each predictor. The LASSO paths are unstable and unsmooth. In contrast, the elastic net has much smoother
solution paths, and the coefficient paths of highly related samples tend to coincide with each other, which clearly shows the group selection effect. Fig. \ref{elasticnetcoef}
presents the reconstruction coefficients of the fifth face image of the first individual using the LASSO model and our elastic net model respectively. The parameter $\lambda$
is set as 0.02 for our model and as 2.6 for LASSO, such that the two models find roughly the same number of non-zero coefficients. A number of highly correlated samples surrounding
the prominent samples are selected in the elastic net, which also demonstrates the group selection effect. However, the prominent samples spread independently in the LASSO model.

\begin{figure}
\begin{center}
   \subfigure[]{\label{subfig:61} \includegraphics[width=.95\linewidth]{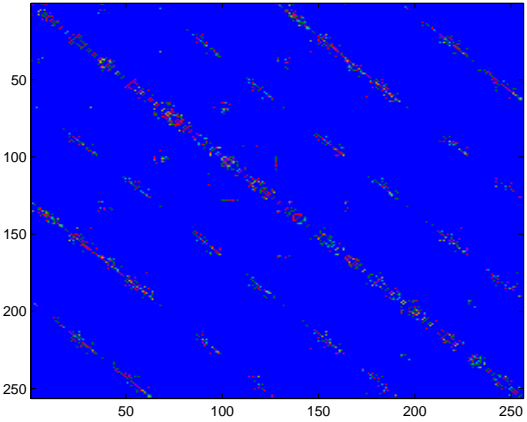}}
   \subfigure[]{\label{subfig:62} \includegraphics[width=.95\linewidth]{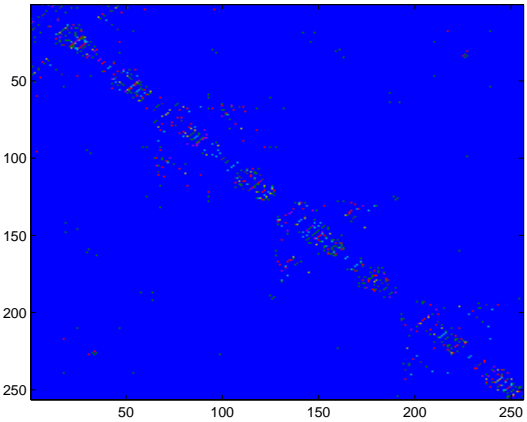}}
   \subfigure[]{\label{subfig:63} \includegraphics[width=.95\linewidth]{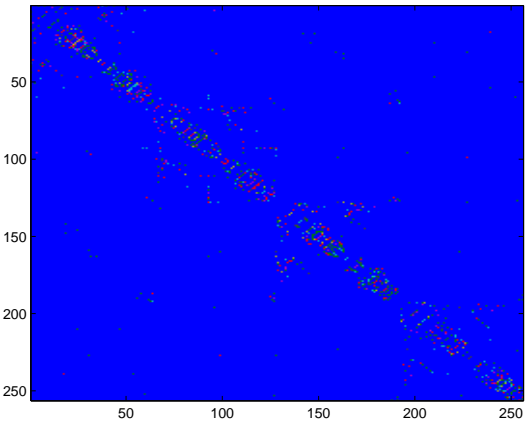}}
\end{center}
\caption{Visualization of coefficient matrixes of different method on the first four individuals of the Extended Yale B face database. \subref{subfig:61} KNN method, \subref{subfig:62} LASSO method and \subref{subfig:63} our method. }
\label{affinitymatrix}
\end{figure}

Fig. \ref{affinitymatrix} depicts the coefficients matrix of KNN, LASSO and our elastic net on the first four individuals of the Extended Yale B face database. The KNN method employs
the Gaussian kernel function to find 45 neighbors of each sample. As with the KNN method, LASSO and our elastic net only keeps the first 45 large coefficients of each sample. The face
samples are arranged sequentially according to their category. Thus, the ideal coefficient matrix should have the block diagonal structure. However, the KNN method has many large
coefficients deviating from the main diagonal. LASSO and our Elastic Net has a distinct diagonal structure nevertheless. The prominent coefficient of
our elastic net method gather more closely along the main diagonal than the LASSO method. It demonstrates that our method is more capable of finding correct neighbors
than the LASSO method.
\begin{figure}
\begin{flushright}
\subfigure[]{\label{subfig:71} \includegraphics[width=1.30\linewidth]{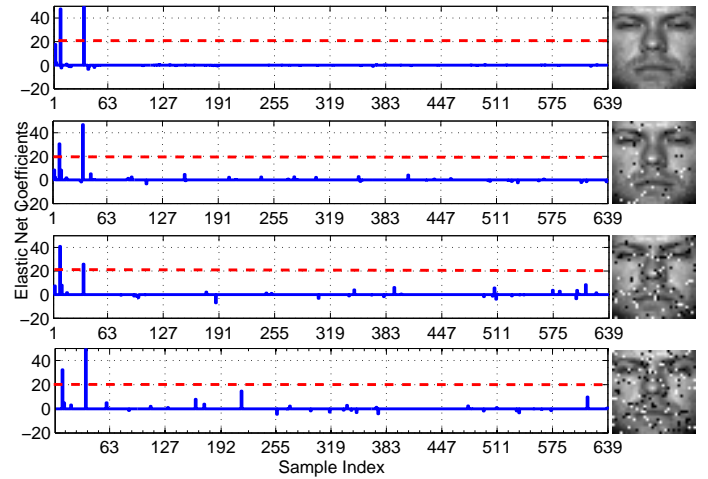}}
\subfigure[]{\label{subfig:72} \includegraphics[width=1.30\linewidth]{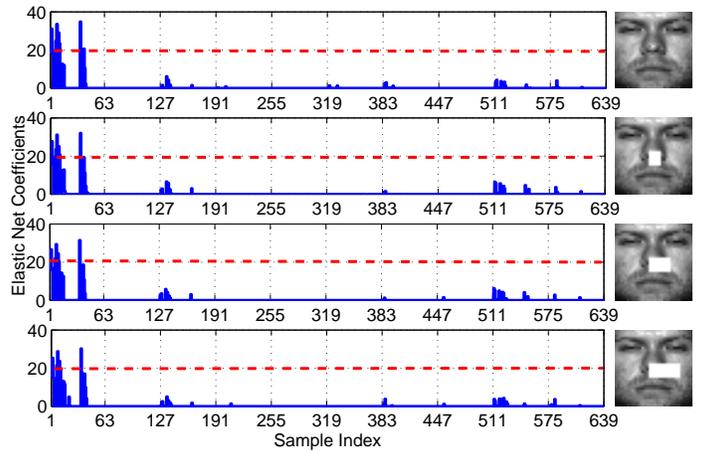}}
\end{flushright}
\caption{Robustness and adaptiveness of hyperedge construction in our elastic net hypergraph.
A sample is used as the response for illustration and the remaining 639 samples from the first
ten individuals are utilized as the dictionary to represent this response sample image. \subref{subfig:71} sparse noise and \subref{subfig:72} data missing.
}
\label{hyperedgerobust}
\end{figure}

\begin{figure}[!t]
\centering
\includegraphics[scale=0.5]{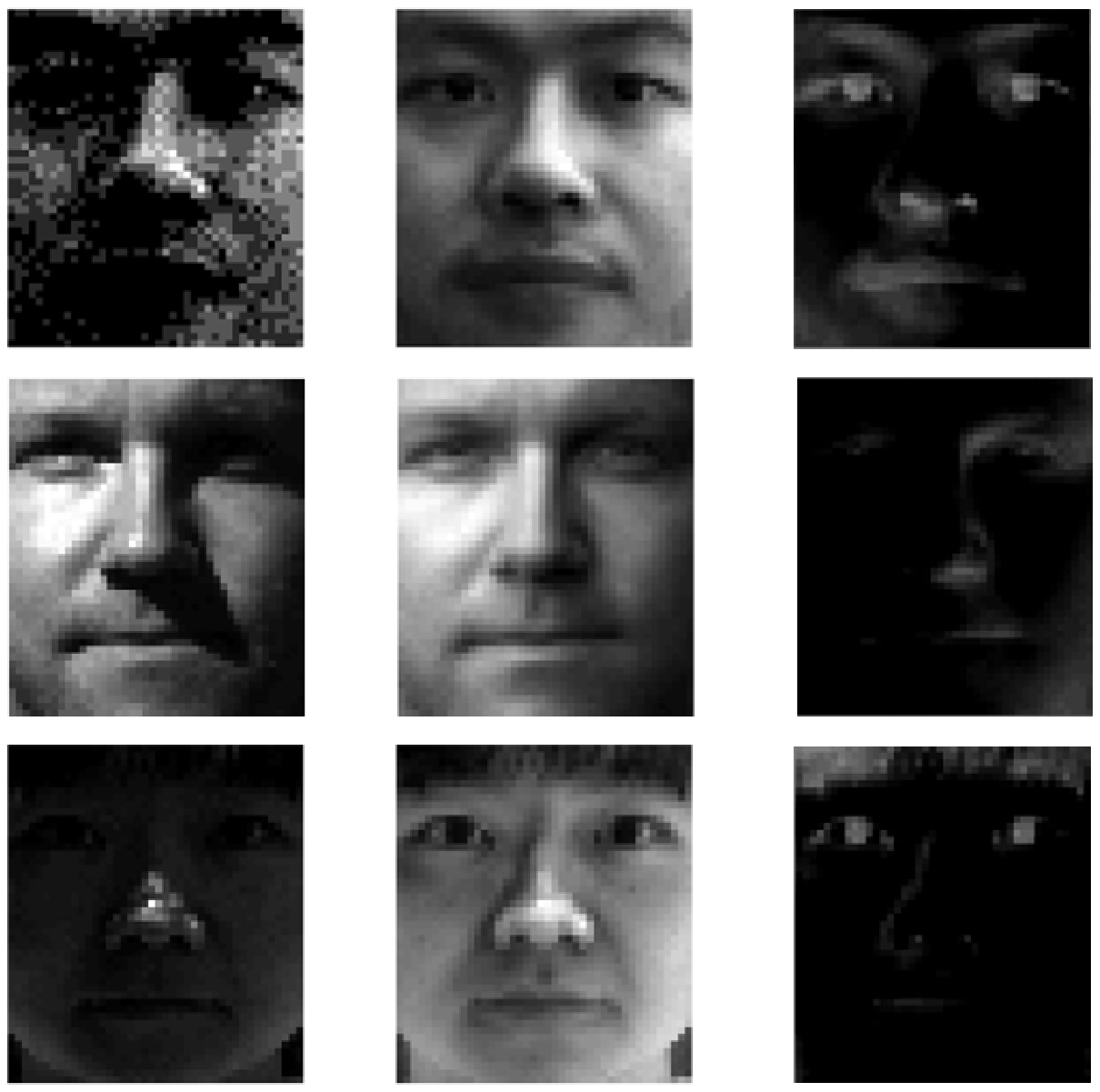}%
\caption{Some examples of using our model to correct the corruptions in faces. Left: The original data; Middle:
The corrected data; Right: The error }
\label{recovery}
\end{figure}

To evaluate the robustness of the hyperedge construction in our elastic net hypergraph, we select the first ten individuals as the sample set. Each individual has 64 samples, thus there are 640 samples in total
and 640 vertices in the constructed hypergraph accordingly. Among the sample set, a sample from the first individual is used as the response for illustration
and the remaining 639 samples are utilized as the dictionary to represent this response sample image. Fig. \ref{hyperedgerobust} shows the results. The horizontal axis indicates the index number of the samples
in the dictionary and the index range is 1 to 639. The vertical axis indicates the distribution of the reconstruction coefficients for the remaining
samples in the elastic net, and the response samples contaminated by the increasing degree of corruption (sparse noise and data missing) are shown in the right column.
Those samples for which the coefficients are beyond the threshold $\theta$ indicated by the red dash line are enclosed by the hyperedge. By this selection strategy, the
number of neighbors, i.e. the size of the hyperedge in ENHG, is adaptive to distinctive neighborhood structure of each datum, which is valuable for applications with non-homogeneous data
distributions. Although the sparse error increases in the response sample, the distribution of the prominent samples in the elastic net does not show significant changes
and the indices of the prominent samples beyond the threshold $\theta$ remain. The main reason for this stability is that the elastic net model can
sperate the error from the corrupted sample. Fig. \ref{recovery} shows the extracted components of some face images. We can see that our model can effectively remove the shadow.
Compared with the hypergraphs constructed by the KNN and $r$-neighborhood methods, the proposed elastic net hypergraph (ENHG) has two inherent advantages. First, ENHG is robust owing to the
elastic net reconstruction from the remaining samples and the explicit consideration of data corruption. Second, the size of each hyperedge is datum-adaptive and automatically determined instead of
uniformly global setting in the KNN and $r$-neighborhood methods.

\section{Experiment Results and Analysis}
We conduct the experiments on three public databases: the Extended Yale face database B \cite{lee2005acquiring}, the PIE face database, and the USPS handwritten digit database \cite{hull1994database},
which are widely used to evaluate clustering and classification algorithms.

\begin{itemize}
    \item \textbf{Extended Yale Face Database B:} This database has 38 individuals, and each subject has approximately 64 near frontal images under different illuminations. Following to \cite{lee2005acquiring}, we crop the images by fixing the eyes and resize them to the size of $32\times32$, and we select the first 10, 15, 20, 30 and full subject set for the respective experiments.
  \item \textbf{PIE Face Database:} This database contains 41368 images of 68 subjects with different poses, illumination and expressions. Similar to \cite{studentgraph}, we select the first 15 and 25 subjects and only use the images of five near frontal poses (C05, C07, C09, C27, C29) under different illuminations and expressions. Each image is cropped and resized to the size of $32\times32$.
  \item \textbf{USPS Handwritten Digital Database:} This database contains ten classes (0-9 digit characters) and 9298 handwritten digit images in total. 200 images are randomly selected from each category for experiments. All of these images are normalized to the size of $16\times16$ pixels.
\end{itemize}

\begin{figure}
\centering
\subfigure[Extended Yale B Sample Images]{\includegraphics[width=3.0in]{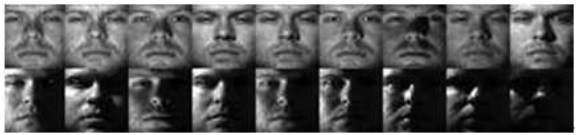}%
\label{Extended Yale B}} \hfil
\subfigure[PIE Sample Images]{\includegraphics[width=3.0in]{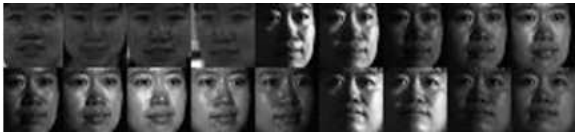}%
\label{PIE}} \hfil
\subfigure[USPS Sample Images]{\includegraphics[width=3.0in]{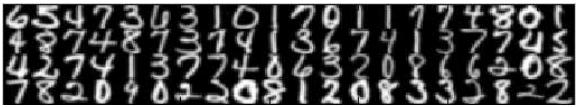}%
\label{USPS}} \caption{Sample images used in our experiments.}
\label{sampleimages}
\end{figure}

Fig. \ref{sampleimages} shows the sample images from the above three databases. As in \cite{wright2010sparse}, we normalize the samples so that they have a unit norm.
To further evaluate the performance of the proposed methods, we compare them to seven state-of-the-art graph-based algorithms including:
\begin{itemize}
  \item \textbf{G-graph:} We adopt Euclidean distance as our similarity measure, and use a Gaussian kernel to compute a weight for each edge of the graph.
  \item \textbf{LE-graph:} Following the example of \cite{belkin2003laplacian}, we construct the LE-graph, which used in Laplacian EigenMaps algorithm.
  \item \textbf{$l_1$-graph: } Following the example of \cite{cheng2010learning}, we construct the $l_{1}$-graph. Since the weight matrix $W$ of a $l_{1}$-graph are asymmetric, we also symmetrize it as suggested in \cite{cheng2010learning}.
  \item \textbf{KNN-hypergraph (KNN-HG):} Following \cite{huang2010image}, \cite{huang2011unsupervised}, we first use the Euclidean distance as the similarity measure. Each sample chooses eight nearest neighbors to construct the hyperedge, then transforms the hypergraph into an induced graph whose edge weights are normalized by the degree of the hyperedge.
  \item \textbf{Semantic correlation hypergraph (SCHG):} Following \cite{hong2012hypergraph}, we construct a semantic correlation hypergraph and each hyperedge is constructed by the index of the top five reconstruction coefficients from the sparse representation, hyperedge weights are then derived by these coefficients.
  \item \textbf{Sparse Subspace Clustering (SSC):} By representing each sample as the sparse combination of all the other data points, spectral clustering is used to obtain the clustering of the data \cite{sparsesubspace}.
  \item \textbf{Low Rank Representation (LRR):} This algorithm \cite{LRR} sought the lowest-rank representation among all the candidates, which is used to define the weighted pairwise graph for spectral clustering.
\end{itemize}

For the sake of evaluating the effect of ENHG, we also implement a $l_1$-Hypergraph algorithm, in which the elastic net is replaced by the original $l_1$ norm constrained sparse representation in the hyperedge construction.

The parameter $\lambda$ and $\gamma$ of the ENHG model are estimated by cross-validation,
and we find that $\lambda$=0.01 and $\gamma$=0.18 is a proper parameter setting. The parameters of all the other algorithms are also tuned for optimal results. All the
algorithms are implemented in Matlab R2011b running on Windows7, with an Intel (R)-Core(TM) i7-2600 3.40GHz processor and 16GB memory. The experiments are run 10 times
and their average results are reported.

\subsection{Spectral clustering experiments}
We carry out the spectral clustering experiments on two face databases and the USPS digital database. Two popular metrics, accuracy (AC) and normalized mutual
information (NMI) \cite{cheng2010learning}, are used for quantitative performance evaluation.

The experimental results are listed in Tables \ref{YaleB clustering}-\ref{USPS clustering} respectively. From the results, it can be seen that the ENHG-based
spectral clustering algorithm achieves better performance than the other five algorithms. The superiority of ENGH is mainly credited to the utilization of the
elastic net to find the overall contextual information for constructing the hyperedge and computing weight. The Hypergraph-based algorithms mostly obtain better accuracy
than the corresponding graph-based algorithms, which shows that the high-order local group information among the data is very useful for clustering. Meanwhile, ENGH
can still obtain good clustering results on the Extended Yale B database with large shadow, which demonstrates its robustness to noise and error in the samples.

\begin{table*}
  \centering
\caption{Comparison of the clustering accuracy (the accuracy/AC and the normalized mutual information/NMI) for spectral clustering algorithms based on ENHG and other methods on the Extended Yale Face Database B.}\label{YaleB clustering}
  \renewcommand{\multirowsetup}{\centering}
        \begin{tabular}{|c|c|c|c|c|c|c|c|c|c|c|c|}
        \hline
        \textbf{YaleB} & \multirow{2}{*}{\textbf{Metric}} & \multirow{2}{*}{\textbf{G-graph}} & \multirow{2}{*}{\textbf{LE-graph}} & \multirow{2}{*}{\textbf{$l_{1}$-graph}} & \multirow{2}{*}{\textbf{SSC}} & \multirow{2}{*}{\textbf{LRR}}&  \multirow{2}{*}{\textbf{KNN-HG}} & \multirow{2}{*}{\textbf{SCHG}} & \multirow{2}{*}{\textbf{$l_1$-Hypergraph}} &\multirow{2}{*}{\textbf{ENHG}}\\
         Cluster\#  &  &  &  &  &  & &  & & & \\
\hline
    \multirow{2}{*}{K=10}     & AC    & 0.172 & 0.420 & 0.758  & 0.821  & 0.822   & 0.507 & 0.775 & 0.873  & \textbf{0.928}  \\
                              & NMI   & 0.091 & 0.453 & 0.738  & 0.811  & 0.814   & 0.495 & 0.702 & 0.846  & \textbf{0.922}  \\
\hline
    \multirow{2}{*}{K=15}     & AC    & 0.136 & 0.464 & 0.762  & 0.801  & 0.816   & 0.494 & 0.791 & 0.896  & \textbf{0.921}   \\
                              & NMI   & 0.080 & 0.494 & 0.759  & 0.767  & 0.802   & 0.464 & 0.749 & 0.866  & \textbf{0.914}   \\
\hline
    \multirow{2}{*}{K=20}     & AC    & 0.113 & 0.478 & 0.793  & 0.797  & 0.801   & 0.534 & 0.782 & 0.884  & \textbf{0.918}  \\
                              & NMI   & 0.080 & 0.492 & 0.786  & 0.781  & 0.792   & 0.485 & 0.742 & 0.866  & \textbf{0.912}   \\
\hline
    \multirow{2}{*}{K=30}     & AC    & 0.08  & 0.459 & 0.821  & 0.819  & 0.807   & 0.512 & 0.773 & 0.876  & \textbf{0.911}   \\
                              & NMI   & 0.090 & 0.507 & 0.803  & 0.814  & 0.806   & 0.484 & 0.737 & 0.856  & \textbf{0.933}   \\
\hline
    \multirow{2}{*}{K=38}     & AC    & 0.08  & 0.443 & 0.785  & 0.794  & 0.785   & 0.486 & 0.764 & 0.826  & \textbf{0.881}    \\
                              & NMI   & 0.110 & 0.497 & 0.776  & 0.787  & 0.781   & 0.473 & 0.723 & 0.804  & \textbf{0.915}    \\
\hline
        \end{tabular}
\end{table*}

\begin{table*}
  \centering
\caption{Comparison of the clustering accuracy (the accuracy/AC and the normalized mutual information/NMI) for spectral clustering algorithms based on ENHG and other methods on the PIE database.}\label{PIE clustering}
  \renewcommand{\multirowsetup}{\centering}
        \begin{tabular}{|c|c|c|c|c|c|c|c|c|c|c|}
        \hline
        \textbf{PIE} & \multirow{2}{*}{\textbf{Metric}} & \multirow{2}{*}{\textbf{G-graph}} & \multirow{2}{*}{\textbf{LE-graph}} & \multirow{2}{*}{\textbf{$l_{1}$-graph}}
        & \multirow{2}{*}{\textbf{SSC}} & \multirow{2}{*}{\textbf{LRR}}& \multirow{2}{*}{\textbf{KNN-HG}} & \multirow{2}{*}{\textbf{SCHG}} & \multirow{2}{*}{\textbf{$l_1$-Hypergraph}} & \multirow{2}{*}{\textbf{ENHG}}\\
         Cluster\#  &  &  &  &  &  & &  & & & \\
\hline
    \multirow{2}{*}{K=15} & AC    & 0.144 & 0.158 & 0.786   & 0.798  & 0.802 & 0.554 & 0.792  &0.801  & \textbf{0.821} \\
                          & NMI   & 0.090 & 0.114 & 0.762   & 0.803  & 0.813 & 0.503 & 0.769  &0.775  & \textbf{0.839} \\
\hline
    \multirow{2}{*}{K=25} & AC    & 0.131 & 0.149 & 0.771   & 0.782  & 0.794 & 0.554 & 0.781  &0.788  & \textbf{0.813} \\
                          & NMI   & 0.087 & 0.106 & 0.753   & 0.766  & 0.760 & 0.503 & 0.763  &0.757  & \textbf{0.828} \\
\hline
        \end{tabular}
\end{table*}

\begin{table*}
\centering
\caption{Comparison of the clustering accuracy (the accuracy/AC and the normalized mutual information/NMI) for spectral clustering algorithms based on ENHG and other methods on the USPS database.}\label{USPS clustering}
  \renewcommand{\multirowsetup}{\centering}
        \begin{tabular}{|c|c|c|c|c|c|c|c|c|c|c|}
        \hline
        \textbf{USPS} & \multirow{2}{*}{\textbf{Metric}} & \multirow{2}{*}{\textbf{G-graph}} & \multirow{2}{*}{\textbf{LE-graph}} & \multirow{2}{*}{\textbf{$l_{1}$-graph}} &
        \multirow{2}{*}{\textbf{SSC}} & \multirow{2}{*}{\textbf{LRR}} & \multirow{2}{*}{\textbf{KNN-HG}} & \multirow{2}{*}{\textbf{SCHG}} & \multirow{2}{*}{\textbf{$l_1$-Hypergraph}} & \multirow{2}{*}{\textbf{ENHG}} \\
         Cluster \#  &  &  &  &  &  & &  & & & \\
\hline
    \multirow{2}{*}{K=4}  & AC    & 0.516 & 0.711 & 0.980 & 0.989  & 0.992 & 0.911 & 0.986   & 0.990            & \textbf{0.996}\\
                          & NMI   & 0.482 & 0.682 & 0.968 & 0.969  & 0.971 & 0.803 & 0.970   & 0.972            & \textbf{0.984}\\
\hline
    \multirow{2}{*}{K=6}  & AC    & 0.424 & 0.69  & 0.928 & 0.936  & 0.957 & 0.871 & 0.925   & 0.945            & \textbf{0.980}\\
                          & NMI   & 0.351 & 0.542 & 0.917 & 0.928  & 0.937 & 0.762 & 0.916   & 0.927            &  \textbf{0.942}\\
\hline
    \multirow{2}{*}{K=8}  & AC    & 0.412 & 0.602 & 0.898 & 0.908  & 0.910 &0.779  & 0.907   & 0.910            & \textbf{0.955}\\
                          & NMI   & 0.252 & 0.503 & 0.905 & 0.894  & 0.903 & 0.641 & 0.882   & 0.910            & \textbf{0.911}\\
\hline
    \multirow{2}{*}{K=10} & AC    & 0.338 & 0.582 & 0.856 & 0.881  & 0.889 & 0.765 & 0.801   & 0.886            & \textbf{0.932}\\
                          & NMI   & 0.213 & 0.489 & 0.872 & 0.866  & 0.871 & 0.636 & 0.822   & 0.870            & \textbf{0.874}\\
\hline
        \end{tabular}
\end{table*}

\begin{table*}
  \centering
  \caption{Classification accuracy rates (\%) of various graphs under different percentages of labeled samples (shown in parenthesis after the dataset name). The bold numbers are the lowest error rates under different sampling percentages.
}\label{classification}
  \begin{tabular}{cccccccc}
    \hline
    \textbf{Dataset} & \textbf{G-graph} & \textbf{LE-graph} & \textbf{$l_{1}$-graph} & \textbf{KNN-HG} & \textbf{SCHG} & \textbf{$l_1$-Hypergraph} & \textbf{ENHG}\\
    \hline
    Extended Yale B (10\%) & 66.49 & 70.79 & 76.34 & 71.80 & 77.68 & 82.15 & \textbf{90.71}\ \\
    Extended Yale B (20\%) & 65.34 & 69.97 & 80.46 & 75.54 & 81.80 & 83.48 & \textbf{92.36}\ \\
    Extended Yale B (30\%) & 33.72 & 71.85 & 81.90 & 77.67 & 82.84 & 85.36 & \textbf{93.94}  \ \\
    Extended Yale B (40\%) & 66.28 & 71.34 & 83.61 & 80.59 & 83.55 & 86.90 & \textbf{94.34}   \ \\
    Extended Yale B (50\%) & 66.90 & 71.60 & 84.75 & 80.80 & 84.48 & 87.08 & \textbf{95.07}   \ \\
    Extended Yale B (60\%) & 67.52 & 71.48 & 88.48 & 81.79 & 89.46 & 90.42 & \textbf{95.28}  \ \\
    \hline
    PIE (10\%) & 65.72 & 67.75 & 78.29 & 68.74 & 79.35 & 80.24 & \textbf{88.32} \ \\
    PIE (20\%) & 66.94 & 69.58 & 82.82 & 70.18 & 84.74 & 84.55 & \textbf{94.93}  \ \\
    PIE (30\%) & 69.89 & 73.48 & 87.94 & 74.39 & 88.78 & 89.29 & \textbf{96.47} \ \\
    PIE (40\%) & 71.54 & 76.38 & 90.99 & 76.14 & 90.33 & 91.75 & \textbf{97.32} \ \\
    PIE (50\%) & 73.04 & 78.35 & 93.39 & 78.76 & 92.66 & 93.71 & \textbf{ 97.65} \ \\
    PIE (60\%) & 74.91 & 80.44 & 95.00 & 79.95 & 94.12 & 94.87 & \textbf{ 98.44}  \ \\
    \hline
    USPS (10\%) & 96.87 & 96.79 & 88.33  & 96.51 & 97.08             & 97.20       &\textbf{97.36}\    \\
    USPS (20\%) & 97.78 & 97.90 & 91.11  & 98.17 & 98.12             & 98.29       &\textbf{98.27}\     \\
    USPS (30\%) & 98.45 & 98.47 & 93.08  & 98.78 & 98.87             & 98.85       &\textbf{98.90}\  \\
    USPS (40\%) & 98.80 & 98.82 & 95.96  & 99.08 & 99.08             & \textbf{99.10}       &99.08      \\
    USPS (50\%) & 99.18 & 99.14 & 97.31  & 99.39 & \textbf{99.41}    & 99.39       &99.40       \\
    USPS (60\%) & 99.35 & 99.28 & 98.86  & 99.51 & 99.50             & 99.52       &\textbf{99.54}\      \\
    \hline
  \end{tabular}
\end{table*}

\subsection{Semi-supervised classification experiments}
We also use the above three databases to evaluate the performance of semi-supervised classification. For the Extended Yale B and PIE databases, we randomly select 50 images from
each subject in each run. The Extended Yale B and the first 15 subjects of PIE are used for evaluation. Of these images, the percentage of the labeled images ranges from 10\% to 60\%.
For the USPS database, the ten digits are used and 200 images are randomly selected from each category for the experiments. These images are randomly labeled with different ratio as in the
face databases. The accuracy rate is used to measure the classification performance as in \cite{cheng2010learning}, \cite{wright2010sparse}, \cite{yan2009semi}. The experimental results are reported in Table \ref{classification}.
We can see that the ENHG method almost always achieves the best classification accuracy compared to the other five methods. The Hypergraph-based methods essentially outperform the pair-wise
graph based methods. $l_1$-Hypergraph has an evident advantage over $l_1$-Graph, which shows that the high-order modeling ability of the hypergraph is very useful for semi-supervised learning.
ENHG outperforming of $l_1$-Hypergraph indicates that the elastic net can represent group structure hidden in the data more effectively. ENHG is also better than SCHG,
because the hyperedges in ENHG are adaptive to local data distribution and the weight computation is more reasonable.

\subsection{Parameters analysis}
\begin{figure}[!t]
\begin{center}
\includegraphics[width=1.1\linewidth]{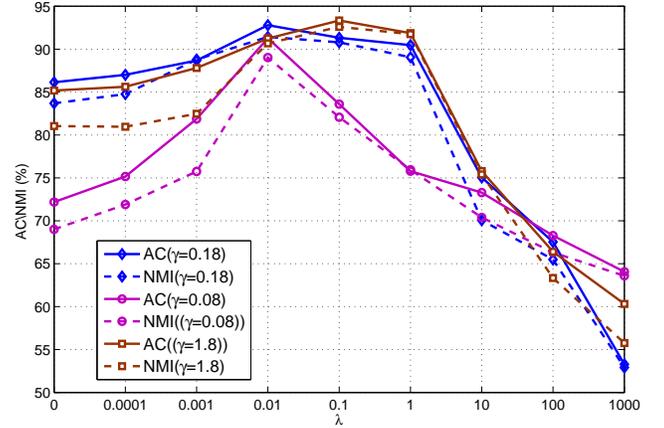}
\end{center}
\caption{Spectral clustering results of our model as a function of $\lambda$ for several values of $\gamma$.}
\label{lambadaclustering}
\end{figure}

\begin{figure}[!t]
\begin{center}
\includegraphics[width=1.1\linewidth]{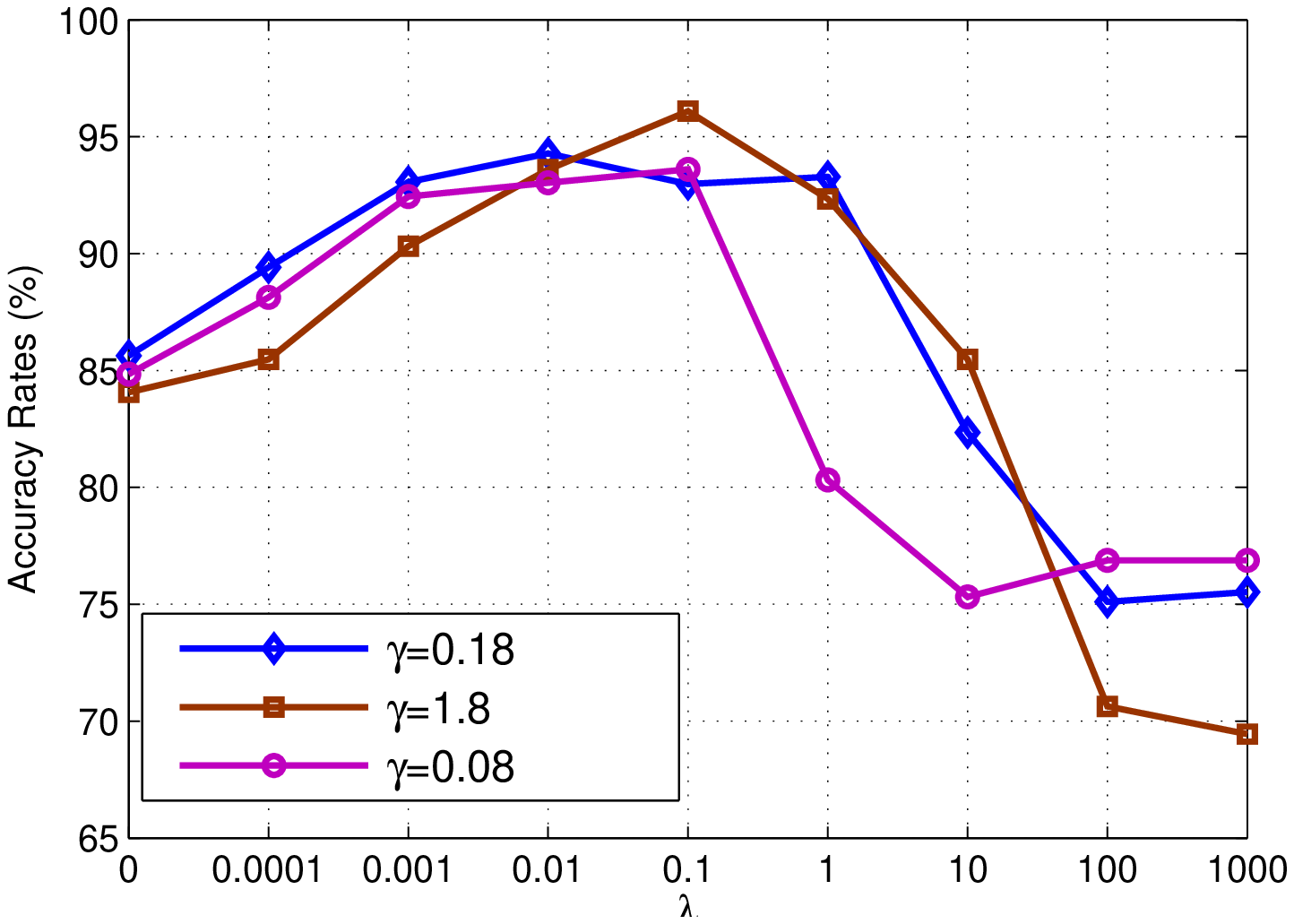}
\end{center}
\caption{Semi-supervised classification accuracy rates of our model as a function of $\lambda$ for several values of $\gamma$.}
\label{lambadasemi}
\end{figure}

In our proposed method, there are two regularization parameters, i.e., $\lambda$ and $\gamma$. $\lambda$ balances the importance between the $l_1$ norm and the $l_2$ norm.
$\gamma$ is the regularization parameter to trade off the proportion between the $XZ$ component and the $S$ component. We design two experiments to evaluate the influence of the two parameters on
the results. We first analyze the influence of $\lambda$. The first ten individuals of the Extended Yale Face Database B are used as the sample set. We fix $\gamma$ as 0.08, 0.18 and 1.8,
and then sample ten points for $\lambda$ in the range [0, 1000] for each value of $\gamma$. The AC and NMI scores of spectral clustering as a function of $\lambda$ for several values of $\gamma$
are plotted in Fig. \ref{lambadaclustering}. The semi-supervised classification results with 30\% labeled samples are presented in Fig. \ref{lambadasemi}. With regard to three values of $\gamma$,
the curves of AC and NMI scores share similar changing trends and the maximum values of different curves are close to each other. When $\lambda$ is set to 0, our model is identical to the LASSO.
With $\lambda$ ranging in [0.001, 1], the score index climbs slowly and stays for a while, which demonstrates the effectiveness of the elastic net regularization. With $\lambda$ increasing to 1000,
our model tends to be closer to ridge regression and thus the score drops rapidly.

\begin{figure}[!t]
\begin{center}
\includegraphics[width=1.1\linewidth]{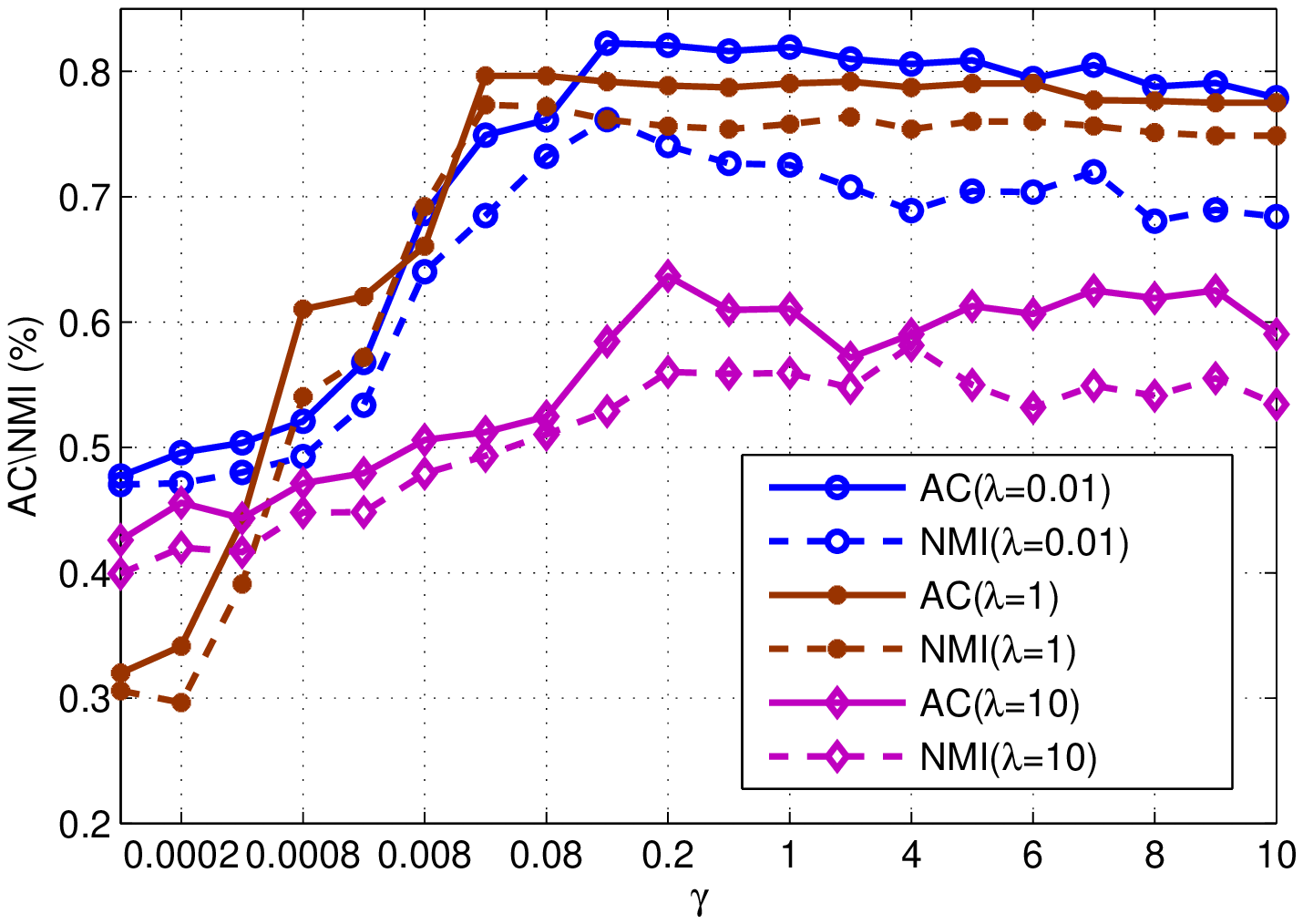}
\end{center}
\caption{Spectral clustering results of our model as a function of $\gamma$ for several values of $\lambda$.}
\label{gammaclustering}
\end{figure}

\begin{figure}[!t]
\begin{center}
\includegraphics[width=1.1\linewidth]{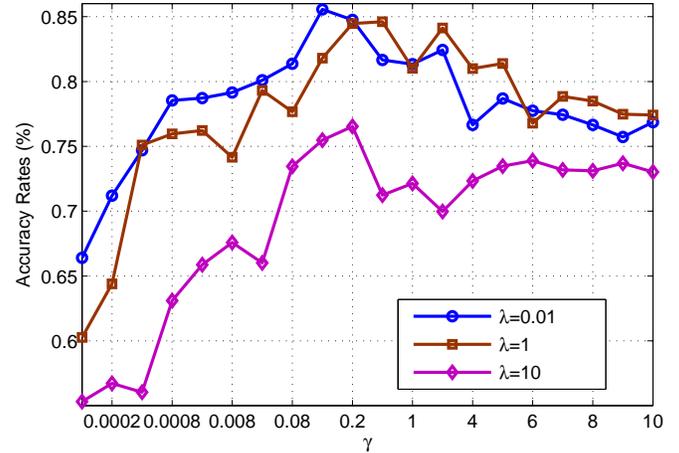}
\end{center}
\caption{Semi-supervised classification accuracy rates of our model as a function of $\gamma$ for several values of $\lambda$.}
\label{gammasemi}
\end{figure}

Furthermore, we turn to the $\gamma$ parameter. The first ten individuals of the Extended Yale Face Database B are also employed as the sample set. Each sample is normalized to have the
unit length. In order to test the influence of the parameter $\gamma$ and validate the robustness of our model to noise, 25\% percentages of samples are randomly chosen to be corrupted by Gaussian noise,
i.e., for a sample vector $x$ chosen to be corrupted, its observed vector is computed by adding Gaussian noise with zero mean and variance 0.1. Fixing $\lambda$ as 0.01, 1 and 10, we run our model
with different $\gamma$ for each value of $\lambda$. Fig. \ref{gammaclustering} plots the spectral clustering results with ENHG. When $\gamma$ is small, the component $XZ$ cannot reconstruct the sample matrix $X$, and $Z$ is
not capable of representing the relationship between samples. Thus, the AC and NMI score are low. As $\gamma$ roughly increases to 0.2, $Z$ can represent the reconstruction relationship effectively,
and noise component may be well separated from $XZ$. The AC and NMI scores reach the top at this time. With $\gamma$ continually increasing to 10, the noise component $S$ cannot be removed well and the AC and NMI scores decrease slowly.
Fig. \ref{gammasemi} presents the semi-supervised classification results, which are similar to the spectral clustering results. However, the changing range of semi-supervised classification is smaller
than spectral clustering. The value of $\lambda$ controls the proportion of $l_2$ norm in the constraint. Although the curves of AC and NMI scores corresponding to each
value of $\lambda$ demonstrate a similar pattern of variability, the maximum scores of each curve are certainly different.


\section{Conclusions}
This paper proposed a novel elastic net hypergraph (ENHG) for two learning tasks, namely spectral clustering and semi-supervised classification,
which has three important properties: adaptive hyperedge construction, reasonable hyperedge weight calculation, and robustness to data noise.
The hypergraph structure and the hyperedge weights are simultaneously derived by solving a problem of robust elastic net representation of the whole data.
Robust elastic net encourages a grouping effect, where strongly correlated samples tend to be simultaneously selected or rejected by the model. The ENHG
represents the high order relationship between one datum and its prominent reconstruction samples by regarding them as a hyperedge. Extensive experiments show
that ENHG is more effective and more suitable than other graphs for many popular graph-based machine learning tasks.

\bibliographystyle{IEEEbib}
\bibliography{IEEEfull,Ref}
\end{document}